\documentclass[10pt, twocolumn, journal,web]{article}
\usepackage{times}
\usepackage[margin=16mm]{geometry}
\usepackage{amsmath,amssymb,amsfonts}
\usepackage{subfig}
\usepackage[utf8]{inputenc}
\usepackage{utfsym}
\usepackage{multirow}
\usepackage[normalem]{ulem}
\usepackage{amsmath,amssymb,amsfonts}
\usepackage{algorithmic}
\useunder{\uline}{\ul}{}
\usepackage{url}
\def\BibTeX{{\rm B\kern-.05em{\sc i\kern-.025em b}\kern-.08em
    T\kern-.1667em\lower.7ex\hbox{E}\kern-.125emX}}
\usepackage{amssymb}
\usepackage{cite}
\usepackage{graphicx}
\usepackage{authblk}

\makeatletter
\renewcommand\AB@affilsepx{, \protect\Affilfont}
\makeatother
\usepackage{tabularx}
\providecommand{\keywords}[1]
{
  \small	
  \textbf{\textit{Keywords---}} #1
}

\begin{document}

\title{\textbf{Lung Infection Severity Prediction Using Transformers with Conditional TransMix Augmentation and Cross-Attention}}
\author[1,3]{Bouthaina Slika}
\author[1, 2, *]{Fadi Dornaika}
\author[4]{Fares Bougourzi}
\author[5]{Karim Hammoudi} 
\affil[1]{\textit{University of the Basque Country}}
\affil[2]{\textit{IKERBASQUE}}
\affil[3]{\textit{Ho Chi Minh City Open University}}
\affil[4]{\textit{ Junia, UMR 8520, CNRS, Central Lille}}
\affil[6]{\textit{Université de Haute-Alsace, IRIMAS}}

\affil[ ]{

\small\texttt{bslika001@ehu.eus, fadi.dornaika@ehu.eus, fares.bougourzi@junia.com, karim.hammoudi@uha.fr}}
\date{}
\maketitle

\date{}
\maketitle
\begin{abstract}
Lung infections, particularly pneumonia, pose serious health risks that can escalate rapidly, especially during pandemics. Accurate AI-based severity prediction from medical imaging is essential to support timely clinical decisions and optimize patient outcomes. In this work, we present a novel method applicable to both CT scans and chest X-rays for assessing lung infection severity. Our contributions are twofold: (i) QCross-Att-PVT, a Transformer-based architecture that integrates parallel encoders, a cross-gated attention mechanism, and a feature aggregator to capture rich multi-scale features; and (ii) Conditional Online TransMix, a custom data augmentation strategy designed to address dataset imbalance by generating mixed-label image patches during training. Evaluated on two benchmark datasets, RALO CXR and Per-COVID-19 CT, our method consistently outperforms several state-of-the-art deep learning models. The results emphasize the critical role of data augmentation and gated attention in improving both robustness and predictive accuracy. This approach offers a reliable, adaptable tool to support clinical diagnosis, disease monitoring, and personalized treatment planning. The source code of this work is available at \url{https://github.com/bouthainas/QCross-Att-PVT}.
\end{abstract}

 \hspace{10pt}

\section{Introduction}
Lung diseases, including chronic obstructive pulmonary disease and interstitial lung disease, pose significant challenges to global health due to their high morbidity and mortality rates \cite{world2020global}. Accurate and timely diagnosis and assessment of these diseases are crucial for effective treatment and management. Traditional diagnostic methods, such as clinical evaluation and pulmonary function tests, often fall short in providing detailed and early insights into disease severity \cite{global2021chronic}.

Medical imaging, particularly chest X-rays (CXRs) and computed tomography (CT) scans, has become indispensable for diagnosing and monitoring lung diseases \cite{adam2020diagnosis, bougourzi2023pdatt}. These imaging modalities offer noninvasive means of visualizing lung pathology, aiding clinicians in their diagnostic and treatment decisions. However, interpreting these images requires significant expertise and is subject to interobserver variability, potentially leading to inconsistent diagnoses and treatment plans \cite{mason2019image, bougourzi2023pdatt}.

In recent years, deep learning has emerged as an advanced technology in medical image analysis, offering automated, accurate, and efficient solutions for various diagnostic tasks \cite{shen2017deep, chen2024TransAttn}. Deep learning models, particularly Convolutional Neural Networks (CNNs), have demonstrated remarkable performance in detecting and classifying lung diseases from CXRs and CT scans \cite{nair2020british, xu2020deep, ghoshal2020estimating}. However, predicting infection severity by quantifying infected regions remains a challenging task due to the limited availability of pixel-wise semantic labels, which are tedious and time-consuming to obtain. Recently, direct approaches for estimating infection severity without the need for pixel-wise labeling have shown promise as a research direction.


Previous studies have highlighted the potential of deep learning models for severity prediction \cite{tang2020quantifying, Santosh2024multimodal}. Despite these advancements, several challenges remain. Lung diseases exhibit a wide spectrum of clinical manifestations, making it difficult for deep learning models to generalize across different patient populations and disease stages \cite{wu2020characteristics}. Furthermore, the limited availability of labeled training data and imbalanced datasets exacerbate issues such as overfitting and bias \cite{wynants2020prediction}. Ensuring the interpretability and generalizability of severity prediction models is also crucial for their clinical adoption \cite{khan2023survey}.

\textcolor{black}{Moreover, in contrast to traditional binary classification approaches, which detect the presence or absence of disease, severity grading via regression provides a more informative and clinically actionable output. Binary models often fail to recognize mild or early-stage findings, as these are typically grouped under the “normal” class \cite{sun2020systematic, bougourzi2024emb}. Such limitations can delay diagnosis and treatment, especially in rapidly progressing conditions like pneumonia. Severity scoring, on the other hand, reflects the continuous nature of disease progression and supports a more nuanced assessment, allowing clinicians to make more individualized decisions regarding treatment intensity, medication dosage, and the appropriate level of care.}


\textcolor{black}{In our previous work presented in \cite{slika2024multi}, we proposed a model that can simultaneously predict multiple scores for quantifying severity using CXRs. However, in this work, we propose a new approach based on a parallel architecture that can adapt to different medical imaging modalities to predict a single score. Our proposed QCross-Att PVT architecture divides the input image into four quadrants, each of which is processed independently by a dedicated backbone. This design allows the model to focus on localized regions, which is particularly important in medical imaging, where regional pathological variations are crucial. The extracted features from each quadrant are then fused via our proposed Cross-Attention Gate, which enables inter-quadrant feature interaction and highlights the most informative regions. Subsequently, a Vision Transformer (ViT) module is employed to capture the long-range dependencies from the aggregated features. Moreover, to address the imbalanced distribution of severity scores across the two datasets, we introduce a Conditional Online TransMix augmentation technique, which dynamically balances the training data and improves model generalization.}

In summary, the main contributions of this work are as follows:

\begin{itemize}
\item Introduction of a novel architecture called QCross-Att-PVT, which utilizes a parallel assembly encoder in conjunction with a transformer and cross-attention to quantify lung disease severity.
\item Evaluation of the proposed method's performance on two medical imaging modalities: Chest X-rays (CXRs) and CT scans.
\item Application of conditional data augmentation techniques to address the training data imbalance problem.
\item Demonstration of an ensemble method that averages the predictions of the best-performing models to enhance performance.
\end{itemize}

The rest of the paper is organized as follows: Section \ref{Relative work} provides an overview of related studies. Section \ref{Proposed methodology} details the proposed generalized pneumonia severity quantification model. Section \ref{Performance Evaluation} presents the performance evaluation, including the datasets used, experimental results, and a comprehensive analysis of each approach in severity assessment. In Section \ref{Discussion}, we discuss the obtained results and interpret the experiments performed in the ablation studies. Finally, Section \ref{conclusion} summarizes the findings and provides concluding remarks.

\section{Related Work}
\label{Relative work}
In recent years, deep learning has made substantial progress in medical image analysis, particularly in diagnosing and evaluating lung diseases through CXRs and CT scans \cite{nafisah2023comparative}. Developing models to predict severity scores has emerged as a crucial research focus, offering quantitative assessments to support clinical decision-making.

\textcolor{black}{For predicting pneumonia infection severity, various deep learning methods have demonstrated their effectiveness. One prominent approach is the use of Convolutional Neural Networks (CNNs). Tang et al. \cite{tang2020quantifying} developed a regression model for assessing pneumonia severity from chest X-rays (CXRs), showing strong correlations with clinical severity scores and supporting early intervention. Similarly, Xu et al. \cite{xu2020deep} introduced a comprehensive framework for detecting and evaluating pneumonia severity using CXRs, emphasizing features closely associated with clinical outcomes.}

\textcolor{black}{In parallel, different CNN architectures have been applied to predict pneumonia infection severity from CT scans by leveraging their feature extraction capabilities. Backbone models such as ResNeSt, ResNet-RS, DenseNet-121, Inception-v3, and various SE-enhanced ResNet variants~\cite{hu_squeeze-and-excitation_2018} have been widely adopted. These CNNs have been used either individually or within ensembles to improve robustness and generalization, such as in the SEnsembleNet approach~\cite{rank2_sensemblenet_2022}. Other strategies enhanced CNN outputs by integrating additional modules like Squeeze-and-Excitation (SE) blocks~\cite{rank2_sensemblenet_2022} or replacing traditional pooling with hybrid pooling mechanisms~\cite{rank3_revitalizing_2022}. Some studies employed contrastive learning on CNN-projected features~\cite{rank5_tricarico_deep_2022}, while others applied mixup data augmentation to improve data diversity and stabilize predictions~\cite{rank6_napoli_spatafora_mixup_2022}. Overall, the trend indicates that modifying or extending standard CNN backbones through ensemble learning, attention mechanisms, and advanced loss formulations significantly enhances COVID-19 infection percentage estimation (CIPE) prediction performance.}

\textcolor{black}{An additional noteworthy approach is proposed by Santosh et al. \cite{Santosh2024multimodal}, who combined multimodal deep learning using both CXRs and CT scans to assess the severity of chronic obstructive pulmonary disease (COPD), thereby improving prediction accuracy by leveraging complementary imaging modalities. Likewise, many researchers have emphasized the crucial role of deep learning in addressing healthcare challenges during the pandemic by enabling efficient lung disease detection and severity quantification~\cite{chen2020deep, PARK2022102299, BOUGOURZI2022, hammoudi2021deep, ajagbe2024deep, kim2015classification}.}

Recent advancements in deep learning, particularly with Vision Transformers (ViTs), have significantly improved the detection and severity quantification of COVID-19 using medical imaging techniques such as CXRs and CT scans. Some of these approaches are discussed below. For instance, Shome et al. introduced the COVID-Transformer, an interpretable model for detecting COVID-19 using Vision Transformers, demonstrating its potential in healthcare applications for accurate diagnostics \cite{shome2021covid}. Kim et al. further explored the capabilities of ViTs in their study, focusing on severity quantification and lesion localization in CXR images, providing an effective tool for assessing disease progression \cite{kim2021severity}. Le Dinh et al. presented a hybrid approach that combined convolutional and transformer neural networks for the classification and severity assessment of COVID-19 from CXR images, showcasing the complementary strengths of these architectures \cite{le2022covid}.

Expanding the use of Vision Transformers beyond X-rays, Al Rahhal et al. developed a model for detecting COVID-19 in both CT and X-ray images, illustrating the versatility of ViTs in handling multiple imaging modalities \cite{al2022covid}. Lentzen et al. applied a transformer-based model to large-scale claims data to predict severe COVID-19 disease progression, demonstrating the broader applicability of these models in clinical decision-support systems \cite{lentzen2023transformer}. Deepa et al. utilized Swin Transformers to identify COVID-19 cases and quantify severity, highlighting the model's precision in detecting different levels of infection \cite{deepa2023swin}.

Furthermore, Taye et al. focused on using ViTs for identifying and classifying the severity of COVID-19 based on thoracic CT images, demonstrating the detailed analysis possible with high-resolution imaging \cite{taye2024thoracic}. Sun et al. advanced this methodology by integrating demographic data with a Swin Transformer model, using enhanced multi-head attention mechanisms to improve diagnostic accuracy \cite{sun2024covid}. Additionally, Huang et al. introduced DeepCoVDR, a deep transfer learning model incorporating graph transformers and cross-attention to predict COVID-19 drug response, highlighting the potential of transformer-based models in pharmacological applications \cite{huang2023deepcovdr}.

The aforementioned studies underscore the growing importance of Vision Transformers in medical image analysis, particularly in the context of COVID-19. In this regard, Nafisah et al. conducted a comparative study between CNNs and Vision Transformers for COVID-19 detection, demonstrating the superior performance of ViTs in capturing complex patterns in medical images, thereby improving detection accuracy \cite{nafisah2023comparative}. Collectively, these studies highlight the benefits of transformer architectures in achieving high accuracy and robustness in various diagnostic tasks, from disease detection to severity assessment and beyond.

Although many studies have employed transformers for diagnostic purposes, most have focused on disease detection or classification. Furthermore, previous studies have typically relied on a single imaging technique for severity evaluation. In contrast, our work aims to quantify lung disease severity by predicting severity scores from either CXRs or CT scans within the same proposed model, ensuring consistent performance.

\textcolor{black}{Cross-attention mechanisms have been extensively studied in recent years \cite{chen2021crossvit, xu2021coat, jaegle2021perceiver, liu2021swin, yuan2021florence}. However, these methods primarily focus on applying cross-attention either between different input patches within the same image \cite{chen2021crossvit, xu2021coat} or across different modalities such as image and text embeddings \cite{jaegle2021perceiver, yuan2021florence}. In contrast, our proposed Cross-Attention Gate (CAG) introduces a novel inter-quadrant attention mechanism specifically designed for medical image severity prediction. Instead of processing the full image as a whole, we partition the image into four quadrants and extract features independently using pretrained PVT backbones. The CAG then performs cross-attention across these quadrants, allowing the model to capture inter-region dependencies and selectively enhance informative regions. This design not only better captures the localized and heterogeneous nature of pathological patterns, but also enables efficient transfer learning from ImageNet pretrained models without requiring full retraining of the attention mechanism.}

\section{Proposed Methodology}
\label{Proposed methodology}
\subsection{Proposed Model}
\begin{figure*}
    \centering
    \includegraphics[
    height=8cm]{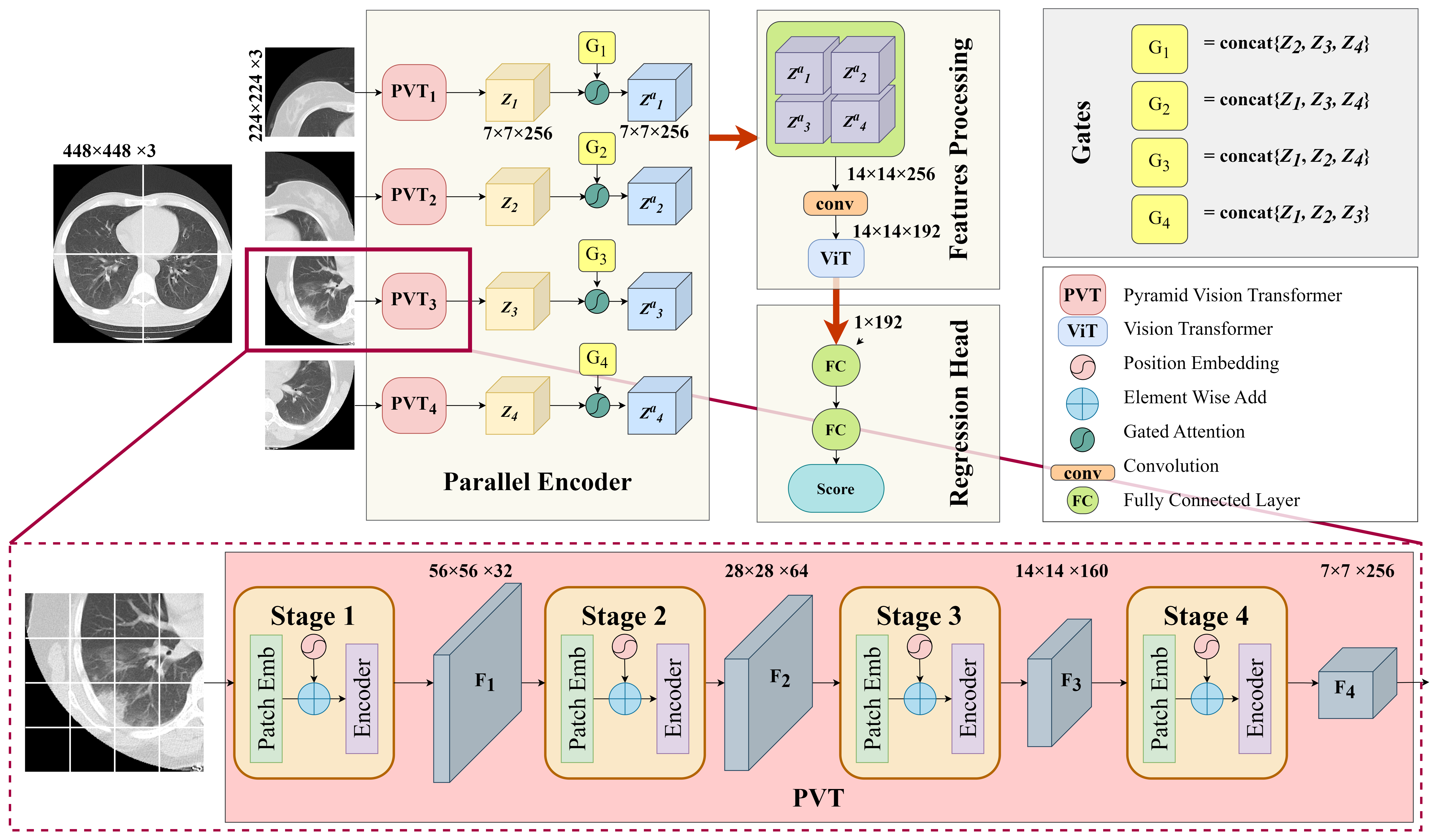}
    \caption{Illustration of the proposed QCross-Att-PVT model.}
    \label{fig: diagram_pvt}
\end{figure*}

In this paper, we propose a new approach for predicting the severity of lung infection in pneumonia, called Quarter Cross-Attention PVT (QCross-Att-PVT), which is evaluated on two medical imaging modalities. As illustrated in Figure \ref{fig: diagram_pvt}, our method consists of three main components: (1) an encoder constructed using parallel transformers with gated cross-attention mechanism, (2) a feature processing module, and (3) a regression head that outputs a scalar value representing the severity of the infection. This design allows the model to simultaneously capture and integrate diverse features from different regions of the input images, improving its ability to extract both local and global information and leading to a more comprehensive assessment of lung disease severity. 

\subsubsection{Encoder}
The input image is partitioned into four equal quadrants, then each is processed independently by a Transformer block. The obtained features from the four quadrants are then processed using our cross-attention gate to capture inter-region dependencies. The encoder is designed with a parallel architecture, where it processes the input through multiple pathways simultaneously.

In more details, the input image $I$ is firstly resized to dimensions of $2H\times2W\times C$ and then divided into four separate regions. Each region, with dimensions $H \times W\times C$, is input into a Pyramid Vision Transformer (PVT)  \cite{wang2021pyramid}. The PVT is an advanced model that adapts the Vision Transformer (ViT) architecture to better handle image recognition tasks, particularly by addressing limitations in capturing multi-scale features. Instead of traditional Multi-Head Attention (MHA), the PVT encoder employs Spatial-Reduction Attention (SRA) to enhance computational efficiency while maintaining strong feature representation. Similar to MHA, SRA processes queries, keys, and values, with significantly reducing the spatial complexity by downsampling keys and values before the attention operation. This reduces memory and computational costs, allowing the model to handle larger input feature maps. The PVT model processes the input through a hierarchical structure with four stages ($k = 1, 2, 3, 4$), where each stage has different embedding dimensions, depths, and attention heads. The output size at each stage is given by:

\begin{equation}
\frac{H}{2^{k-1}\cdot P} \times \frac{W}{2^{k-1}\cdot P} \times {C_k},
\end{equation}

where $P$ is the patch size and $C_k$ is the number of channels after the $k^{th}$ stage. At each stage, the resolution is further downsized with each reduction capturing increasingly broader contextual information as shown in Figure \ref{fig: diagram_pvt}. The final output of each PVT typically has a small spatial resolution with a deeper representation, providing a rich representation that combines global context with fine-grained details.

\begin{figure}
    \centering
    \includegraphics[width= \linewidth]{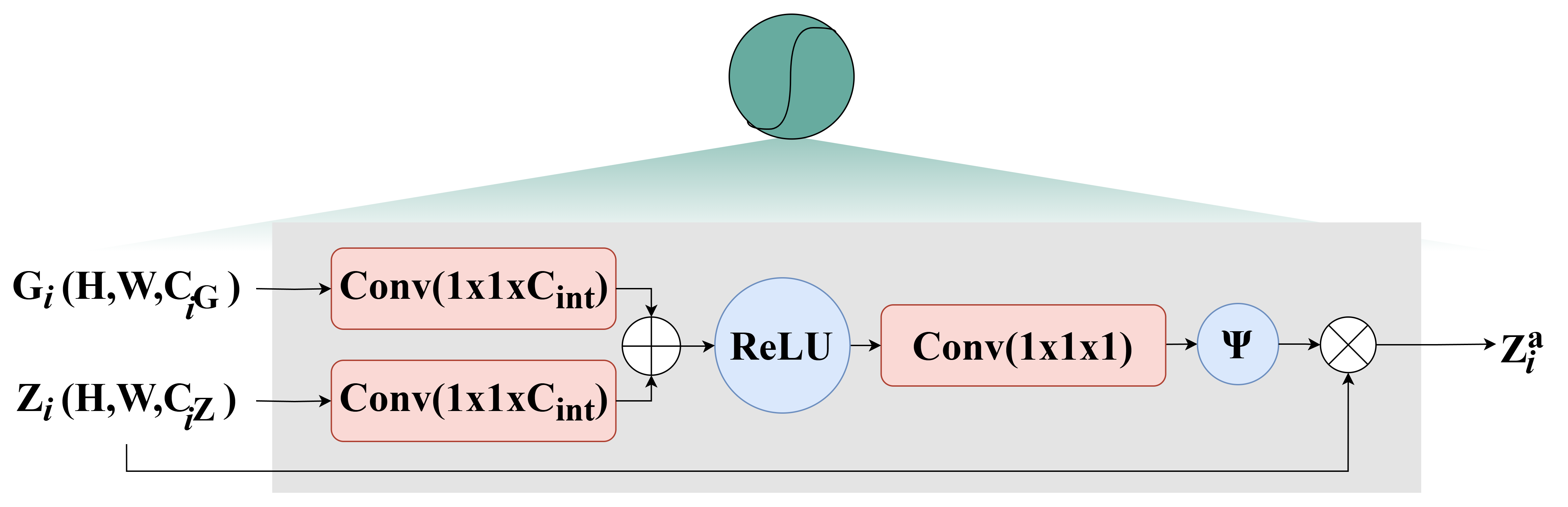}
    \caption{Gated Attention.}
    \label{fig:gated_attention}
\end{figure}

In our setup, $H=224$, $W=224$, $P = 4$, and $ \{C_1, C_2, C_3, C_4 \} = \{ 32, 64, 160, 256\}$. The channel size $160$ in the third stage reflects an experimentally optimized balance between feature richness and computational efficiency, particularly when used with corresponding multi-head attention settings $(1, 2, 5, 8)$, which aligns with the scale of these channels.
The input image $I$ is divided into four sections, resulting in four 3D feature tensors $Z_i$. Each of these tensors is then processed using an cross-Attention Gate (CAG). As depicted in Figure \ref{fig:gated_attention}, the CAG is defined by the following expression:

\begin{eqnarray}
Z^a_i & = & AG\{Z_i,G_i\} \\
& = &  \psi( conv [ ReLU(conv(G_i)+conv(Z_i))]) \otimes Z_i \nonumber 
\end{eqnarray}

In this equation, the inner convolutional layers perform linear transformations using $1 \times 1$ convolutional blocks, which adjust the number of channels in the input and gating signals $C_{Z_i}$ and $C_{G_i}$ to intermediate feature representations with a new channel count $C_{\text{int}}$ for $Z$ and $G$, respectively. The features derived from $C_{Z_i}$ and $C_{G_i}$ are combined and passed through a third $1 \times 1$ convolutional layer, producing a 2D weight map. The sigmoidal activation function $\psi$ is then used to learn the spatial attention coefficients for each patch token. These spatial coefficients are applied to the encoder's feature maps $Z_i$,  $\otimes$ representing element-wise multiplication.

For each set of extracted features from one-quarter of the input image $I$, the gating signal is formed by concatenating the features extracted from the other three quarters, while the features from the current quarter are used as the input signal. The output features for the first quarter are determined by the following equation:

\begin{equation}
\label{eq:att}
Z^a_i = AG\{Z_i,G_i\}, 
\end{equation}

where $Z^a_i$ is the output produced by applying the attention gate to $Z_i$, the $i^{th}$ tensor, where $Z_i$ is the input signal and $G_i$ acts as the gating signal. In this context, for each $i$, $G_i$ is defined as the concatenation of the $3$ other output tensors of the other three tensors. i.e, $G_1 = concat\{Z_2, Z_3, Z_4 \}$, $G_2 = concat\{Z_1, Z_3, Z_4\}$, $G_3 = concat\{Z_1, Z_2, Z_4\}$, and $G_4 = concat\{Z_1, Z_2, Z_3\}$.  
The operation performs cross-attention between the tokens of the current image region and those of the other three regions. This involves calculating attention scores between tokens from different regions, allowing the model to assess the relative importance of tokens across various parts of the image. 

\subsubsection{Features Processing and Regression Head}
The resulting tensors from the four quadrants are  concatenated along the spatial dimension. Specifically, the four tensors resulting from the four partial images are grouped horizontally and vertically, as shown in Figure \ref{fig: diagram_pvt}. A convolution layer is then applied, followed by an additional Vision Transformer (ViT). This additional ViT is used to process the combined features and serves as a feature aggregator. It further refines the information extracted from the different regions before the final prediction of the severity is performed. Finally,  a regression head consisting of two fully connected layers that map the cls features of the ViT into a single scalar representing the predicted severity score, as shown in Figure \ref{fig: diagram_pvt}.

\subsection{Data Augmentation: Conditional Online TransMix}
\label{transMix}

\begin{figure}
    \centering
    \includegraphics[width=1\linewidth]{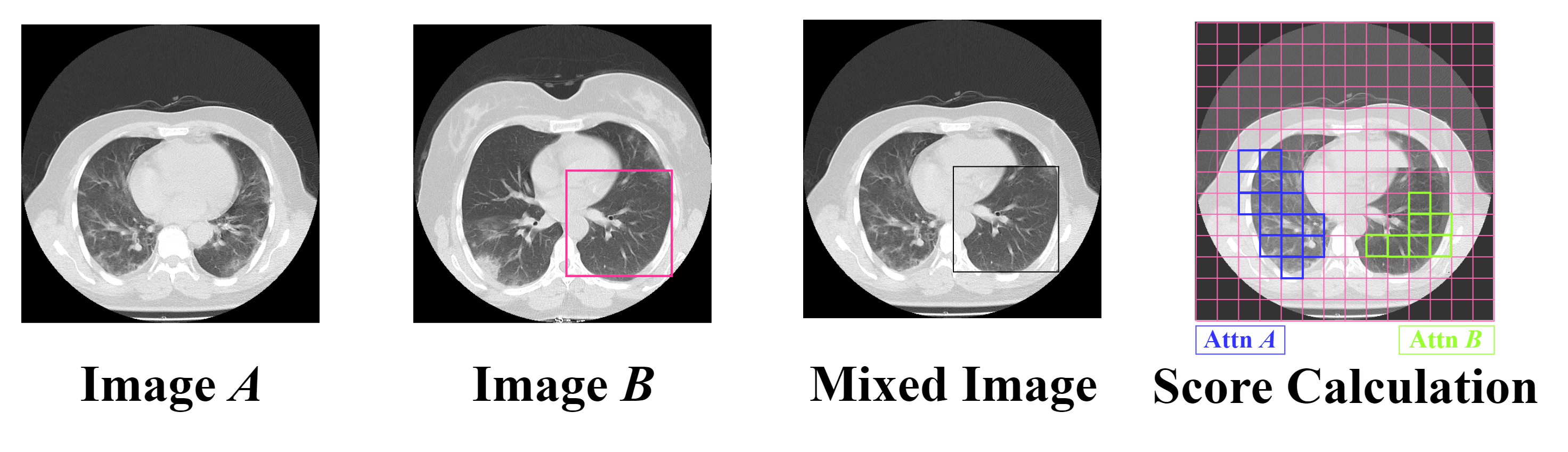}
    \caption{TransMix applied to CT images.}
    \label{fig: transMix}
\end{figure}

In our work, we have addressed the problem of score imbalance in pneumonia severity infection datasets by using an online augmentation technique based on the TransMix method \cite{transmix}. Although this technique was originally proposed for classification tasks, we have adapted it for regression by generating new mixed scores for the newly mixed images. Our novel adaptation of this data augmentation technique leverages attention maps to drive the score mixing process. The mixed image is obtained using the traditional CutMix method \cite{yun2019cutmix}, where a random patch of image $B$ is inserted into image $A$ to form a new mixed image, as shown in Figure \ref{fig: transMix}. After creating the mixed image, its relative ground truth score is calculated following the TransMix approach. This method assigns mixed labels based on a multi-head class attention map computed using self-attention mechanisms similar to those used in ViTs. The class attention map of the augmented image represents the attention of the class token (CLS) to the input patches of the mixed image and highlights the most useful patches of the input image for the final prediction.

TransMix uses this attention map $Att$ to control the process of mixing scores. In this context, $\lambda$ (representing the proportion of the source image) is set to the sum of the weights in the attention map $Att$ that overlaps with the clipping mask, each weight corresponding to the importance assigned to a particular region of the image.

Given images $A$ and $B$, we update the mixing coefficients $\lambda$ using the attention map of the mixed image $Att$ :
\begin{equation}
\lambda = Att \cdot\downarrow (M),
\end{equation}
where $\cdot\downarrow$ represents nearest-neighbor interpolation that downsamples the mask $M$ from $H\times W$ into $P$ pixels. 

 In this way, we can dynamically reassign the weights of the scores depending on how much attention map is directed to each patch. The patches that receive more attention have a higher proportion of the \cite{transmix}.
The ground truth score of the mixed image $\overline{y}$ is:

\begin{equation}
\overline{y} = \lambda*y_B + (1-\lambda)*y_A,\label{eq5}
\end{equation}
where $y_A$ and $y_B$ are the ground truth scores of images $A$ and $B$ respectively.

In our implementation, TransMix was applied conditionally to address dataset imbalance by focusing on underrepresented scores. We specifically performed TransMix augmentation on images with these scores, ensuring the augmented samples enhanced the representation of the low-distribution regions. The decision to apply augmentation was based on a threshold derived from the training dataset's score distribution. By defining a threshold for the ground truth score of the image, the network effectively learned to generate new images and their corresponding labels, improving its ability to generalize from a more balanced and diverse set of training examples. This approach led to better generalization and more accurate severity predictions.

\subsection{Loss Function}
\label{loss}
In our approach, where we train on imbalanced datasets, a user-defined weighted loss function can be helpful to balance the data distribution and ensure that the model focuses on critical features. With this technique, the loss contributions of the different scores are weighted differently. This is particularly useful when images with a certain severity are underrepresented in the training dataset. By weighting the loss associated with these minority categories higher, the model is incentivized to pay more attention to these critical examples during training. The weights are calculated using equation (\ref{eq1}).

\begin{equation}
w_l =\frac{N}{c_l\cdot k}
\label{eq1}
\end{equation} 

Here, $w_l$ is the weight for the $l^{th}$ score level, $N$ is the total number of images in the training dataset, $c_l$ is the number of samples in the $l^{th}$ score level, and $k$ is the total number of score levels.
Since we are dealing with two dataset modalities and their respective scores, in each case, we have different parameter values. For the RALO dataset ($k = 17$), the scores range from $0-8$ with an increment of $0.5$ resulting in $17$ score levels. The Per-COVID-19 CT dataset is labeled with a range of $0-100$ with an increment of $1$ forming $101$ score levels ($k = 101$).
We use these weights to calculate the weighted loss that will be the loss function used while training our model. The weighted loss $\cal{L_W}$ is calculated as follows:
\begin{equation}
{\cal{L_W}} = \sum_{i=1}^{N} w_i |y_i - \hat{y}_i|,
\label{eq2}
\end{equation}
where $w_i $ is the weight for the $i^{th}$ image (computed from Eq. (\ref{eq1})), $ y_i $ is the ground true value for the $i^{th}$ image, $ \hat{y}_i $ is the predicted value for the $i^{th}$ image, and $N $ is the total number of images.
In the context of predicting severity scores for lung diseases from chest images, a custom weighted loss function ensures that the model adequately learns to distinguish between different levels of disease severity, thereby providing more reliable and fine assessments that are crucial for effective patient management.

\section{Performance Evaluation}
\label{Performance Evaluation}
\subsection{Datasets}
\label{dataset}

\subsubsection{RALO Dataset}
\begin{figure}
    \centering
    \subfloat[Distribution of the GE score.]{
    \includegraphics[width=0.9\linewidth]{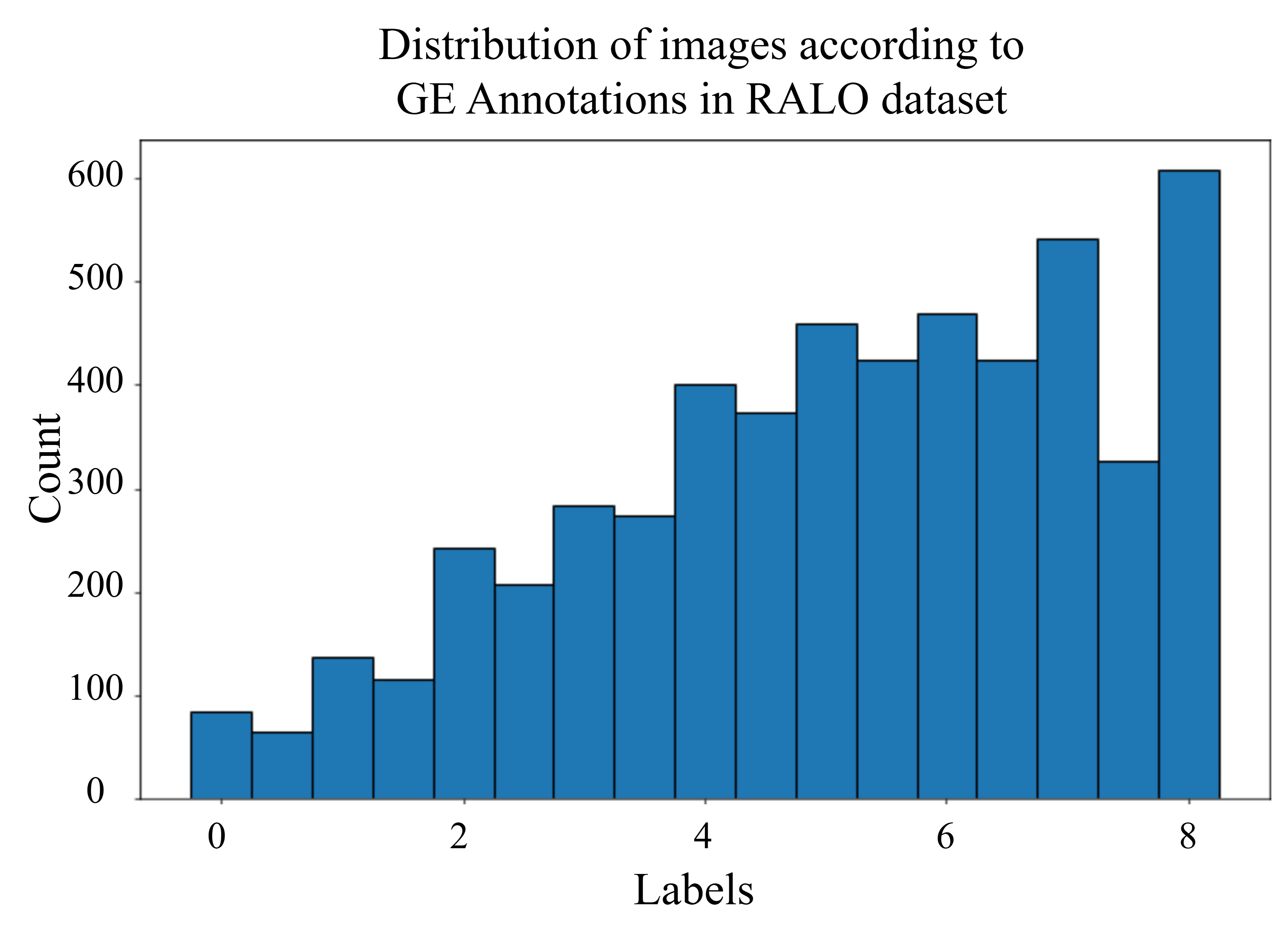}}\\ 
    \subfloat[Distribution for the LO score.]{
     \includegraphics[width=0.9\linewidth]{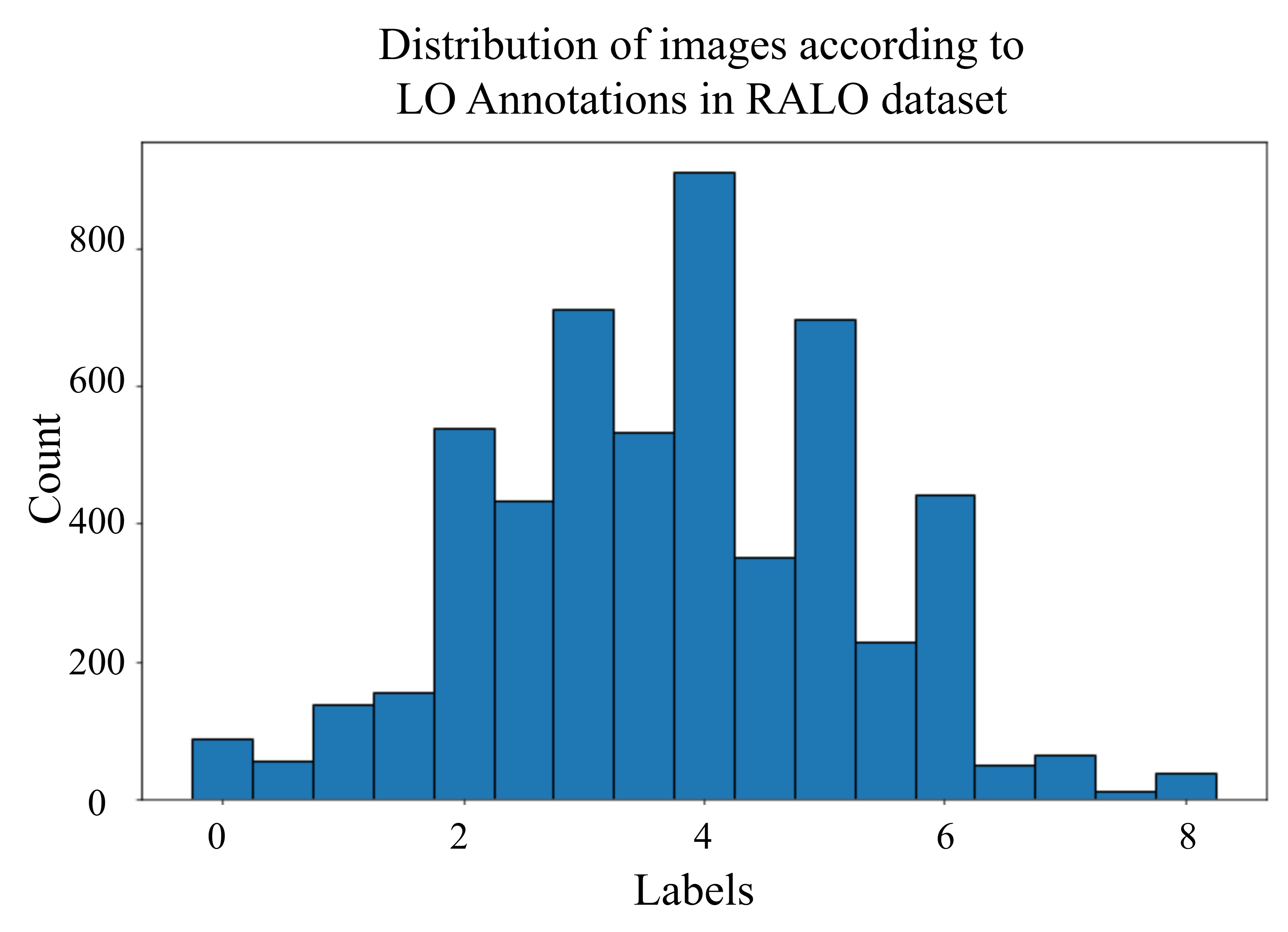}}
    \caption{RALO training set scores distribution.}
    \label{cxr_hist}
\end{figure}

The primary goal of this research is to evaluate the effectiveness of deep learning models in determining the severity of lung diseases. To accomplish this, we utilized the Radiographic Assessment of Lung Opacity Score (RALO) dataset, which consists of 2,373 chest X-ray images \cite{josephpaulcohen20214634000}. These images were carefully evaluated and scored by two expert radiologists from Stony Brook Medicine to create a detailed COVID-19 dataset for research purposes. The dataset is divided into a training set of 1,878 images and a test set of 495 images. In this study, radiological grading focuses on two key criteria: Geographic Extent (GE) and Lung Opacity (LO). GE refers to the spread of lung involvement by ground-glass opacity or consolidation, with separate scores for the right and left lungs. The GE score ranges from 0 (no consolidation) to 4 (maximum consolidation), and the overall GE score is the sum of the left and right lung scores. LO is assessed independently for each lung, with scores ranging from 0 (no opacity) to 4 (total whiteout), reflecting varying degrees of lung opacity. The total LO score, which ranges from 0 to 8 points, is calculated by summing the scores of both lungs. The final ground-truth scores are averages of the two radiologists' evaluations, and they are represented in the set ${0, 0.5, 1.0, 1.5, 2.0, \dots, 8.0}$ \cite{wong2021towards}.  An offline combined lung and score replacement is applied to the training set as done in a previous work \cite{slika2023automatic}. The resultant dataset is distributed as shown in Figure \ref{cxr_hist}.

\subsubsection{Per-COVID-19 Dataset}
\begin{figure}
    \centering
    \includegraphics[width=0.9\linewidth]{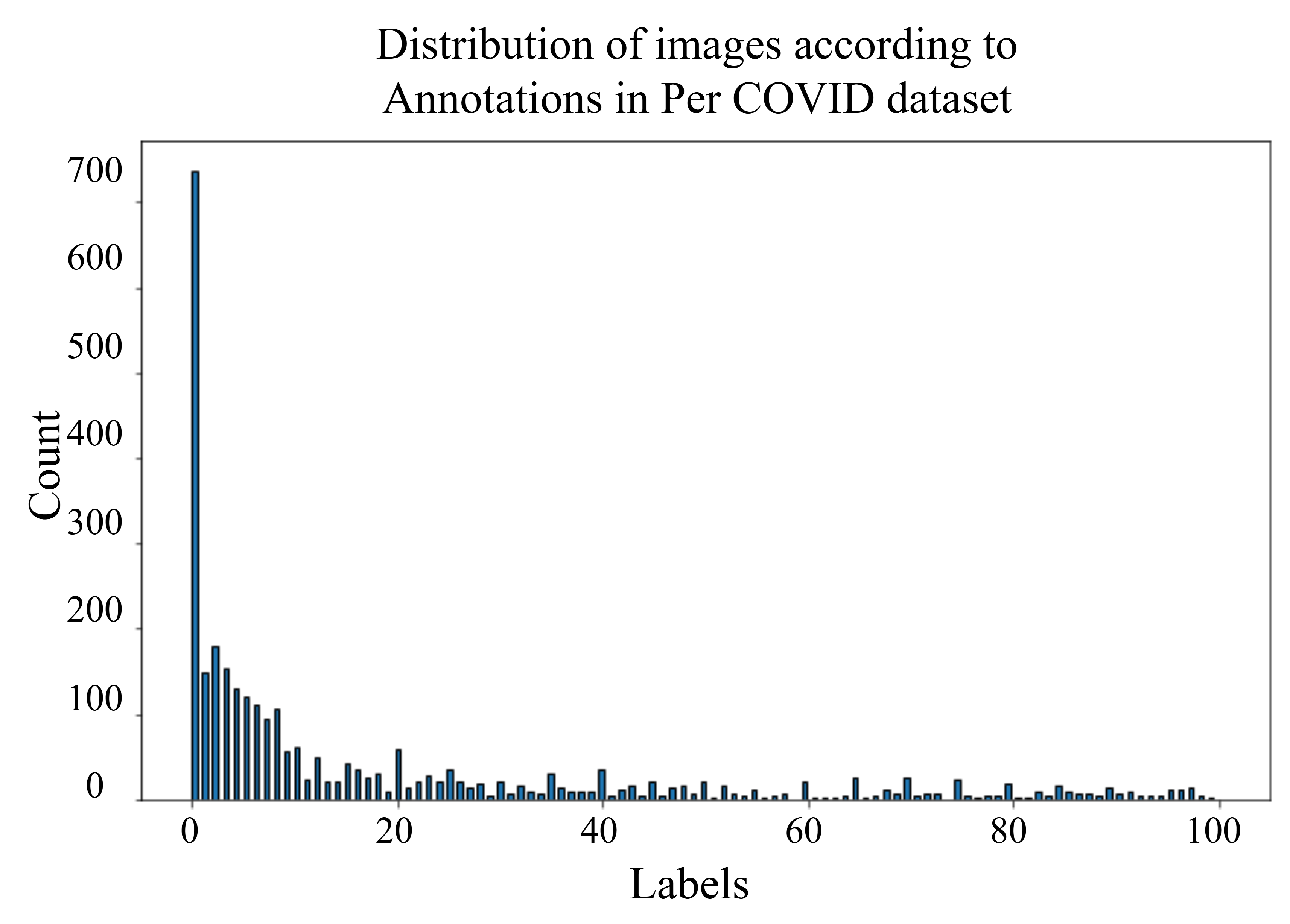}
    \caption{Per-COVID-19 training set with CIP score distribution.}
    \label{ct_hist}
\end{figure}

The Per-COVID-19 dataset's training and validation splits were derived from 189 CT scans, each confirming a COVID-19 infection \cite{bougourzi2021per, vantaggiato2021covid}. The patient demographics include both males and females, aged between 27 and 70 years. Importantly, each CT scan in the Per-COVID-19 dataset corresponds to a single patient, ensuring that the number of scans matches the number of patients. COVID-19 diagnosis in this dataset is based on a positive reverse transcription polymerase chain reaction (RT-PCR) test and findings from CT scans, which two experienced thoracic radiologists carefully interpreted. Each CT scan comprises 40–70 slices, and the radiologists determined the percentage of lung area affected by COVID-19 infection relative to the total lung area. These COVID-19 Infection Percentage (CIP) annotations are expressed as percentages between 0\% and 100\%. The dataset is divided into 3,054 training slices and 1,301 validation slices \cite{bougourzi2021per, vantaggiato2021covid}. The testing split of Per-COVID-19 combines three different COVID-19 segmentation datasets \cite{bougourzi2021per}. For the test data, the ground truth for CIP is calculated by determining the proportion of infected pixels relative to the total number of lung pixels, using both the infection and lung segmentation masks. Producing accurate CIP compared with the ones of training and validation data. The Per-COVID-19 dataset presents additional challenges beyond typical CIP estimation from CT scans. Specifically, the challenge arises from training models with noisy labeled data, as the CIP ground truth for both the training and validation sets was estimated by radiologists on a scale of 100. Additionally, the research addresses domain adaptation issues, as the testing data originates from three sources different from those of the training data \cite{bougourzi2024covid}. Figure \ref{ct_hist} shows the distribution of the training data used in our paper.

\subsection{Experimental Settings}
\textcolor{black}{In our work, we used a weighted loss function (explained in Section \ref{loss}) to address label imbalance in the severity scores, ensuring that underrepresented severity levels contributed proportionally to the total loss during training. For optimization, we employed the AdamW optimizer with a base learning rate set to $10^{-5}$, chosen to provide stable convergence while effectively managing weight decay, an important factor for Transformer-based architectures with a relatively large parameter count (28 million in our case). The AdamW optimizer was configured with $\beta = 0.5$ and $\beta = 0.99$ to balance momentum and adaptivity, settings we found empirically well-suited for our regression tasks.}
To further enhance training stability and model generalization, we integrated a cosine annealing warm restarts scheduler with an initial restart period equal to the total number of iterations per epoch and a multiplication factor of 2. This scheduler progressively reduces the learning rate following a cosine decay pattern and periodically restarts to a higher value, which helps the model escape local minima and explore alternative optima during training. The model was trained for 50 epochs with a batch size of 16 using $448 \times 448$ input images. All experiments were conducted on a Titan X GPU (NVIDIA, driver version 525.125.06) with CUDA version 12.0. 

\subsection{Experimental Results and Comparison}
To assess the severity of pulmonary illness, we applied our model to both CXR and CT modalities.
The refined RALO dataset, which includes 5,634 images, provides GE and LO scores ranging from 0 to 8, indicating the extent of the illness from mild to critical. In our study, we utilized both the original images and those with applied combined lung and score replacement augmentation from a previous work \cite{slika2023automatic}, but with the addition of a conditional online score-correlated TransMix  augmentation strategy. The details of the applied method are described in Section \ref{transMix}.

In this context, the threshold values used for applying Conditional TransMix are chosen according to the initial distribution of the labels across the dataset. Figure \ref{cxr_hist}  illustrates the distribution of GE scores across the dataset, revealing that images with a score greater than 4 are more prevalent than those with a score of 4 or less. This imbalance suggests a natural division within the data, prompting us to use a score of 4 as a threshold. By setting this threshold, we effectively distinguish between cases with higher and lower severity, allowing for targeting the less frequent data with the proposed augmentation method. Similarly, for the LO scores, images with scores below 2 and above 6 are less frequent compared to those within the mid-range scores. This uneven distribution highlights the scarcity of cases with either very low or very high severity, which could pose challenges for the model's training and evaluation. Understanding this distribution is crucial for applying the augmentation to the scarce data to ensure balanced representation and accurate predictions across all severity levels. 

On the other hand, Figure \ref{ct_hist} shows the distribution of scores for CT scans within the range of 0 to 100, revealing that images with a score greater than 10 are less frequent. This indicates that the majority of the dataset consists of images with a score of 0, while those with non-zero scores are comparatively rare. This imbalance highlights the need for careful consideration during model training to ensure that cases with non-zero scores are adequately represented for reliable performance across the full range of scores which is achieved by data augmentation.
Moreover, to consider the generalizability of our model in assessing lung disease severity, we tested its performance on CT images. Using the Per-COVID-19 dataset, which comprises 3,054 images, we trained our proposed model to predict the severity score represented by a scalar that ranges from 0 to 100. During training, conditional online score-correlated TransMix  augmentation was applied to balance the dataset according to the strategy stated in Section \ref{transMix}. 

We have performed comparisons with state-of-the-art architectures for both modalities to emphasize the superior performance of the proposed model. For the CXR modality testing, several deep-learning models are included in this comparison, such as COVID-NET \cite{wong2021towards}, COVID-NET S \cite{wong2020covid}, ResNet50 \cite{he2016deep}, Swin Transformer \cite{liu2021swin}, XceptionNet \cite{chollet2017xception}, Feature Extraction \cite{cohen2020predicting}, MobileNetV3 \cite{mobilenetv3}, InceptionNet \cite{szegedy2015going}, ViTReg-IP \cite{slika2024lung}, MViTReg-IP \cite{slika2024multi}, in addition to our proposed model. Table \ref{tab: comparison CXR} summarizes the key performance metrics.

\begin{table*}
    \centering
    \caption{ \textcolor{black}{Comprehensive Performance Evaluation of the Proposed Method vs. State-of-the-Art Techniques on the RALO CXR Dataset.}}
   \scalebox{0.9}{ \begin{tabular}{|l|c|c|c|c|c|c|c|c|}
            \hline
            & \multicolumn{3}{|c|}{\textbf{GE}} & \multicolumn{3}{|c|}{\textbf{LO}} &\textbf{Number of} &\textbf{Inference}\\ \hline \hline
              \textbf{Model}&\textbf{MAE $\downarrow$}&\textbf{PC $\uparrow$} & \textbf{AE-SD} &\textbf{MAE $\downarrow$}&\textbf{PC $\uparrow$}&\textbf{AE-SD}&\textbf{parameters} &\textbf{time}\\
            \hline   \hline
            \textit{COVID-NET}\cite{wong2021towards}&4.458&0.549 &1.921&2.242&0.535&1.689&12M &$\sim$ 18 ms\\ \hline 
            \textit{COVID-NET-S}\cite{wong2020covid}&4.698&0.591 &2.21&2.254&0.529& 1.899&12M &$\sim$ 17ms\\ \hline 
            \textit{ResNet50} \cite{he2016deep}& 1.094&0.688 &0.682& 1.061& 0.431&0.9723&23M &$\sim$ 26 ms\\ \hline 
            \textit{Swin Transformer}\cite{liu2021swin}&0.916& 0.817 &0.4841&0.803 & 0.697&0.4374&29M  &$\sim$ 40 ms\\ \hline 
            \textit{XceptionNet}\cite{chollet2017xception}&0.854&0.821 &0.3321&0.768 &0.701&0.3011&23M &$\sim$ 32 ms\\ \hline 
            \textit{Feature Extraction}\cite{cohen2020predicting}& 0.967 & 0.753 &0.3912&0.865 & 0.711&0.3766& 2.M &$\sim$ 7 ms\\ \hline 
           \textit{MobileNetV3}\cite{mobilenetv3} &0.847&0.827 &0.2615&0.732&0.738 &0.2398&4.2M &$\sim$ 10 ms\\ \hline 
           \textit{InceptionNet}\cite{szegedy2015going}  & 0.702 &0.886  &0.1894&0.609 &0.829 &0.1655&24M &$\sim$ 29 ms\\ \hline 
           \textit{ViTReg-IP}\cite{slika2024lung}&0.565&0.925 &0.0745& 0.510&0.857 &0.7211&5.5M & $\sim$ 22 ms\\ \hline 
           \textit{MViTReg-IP} \cite{slika2024multi}&0.531&0.938 &0.0651&0.462&0.881&0.0634&11.2M &$\sim$ 24 ms\\[0.5ex] \hline
 \textbf{QCross-Att-PVT}& \textbf{0.362}& \textbf{0.971} &\textbf{0.0453}& \textbf{0.337}& \textbf{0.945}&\textbf{0.0478}&\textbf{28M} &\textbf{$\sim$ 45 ms}\\\hline
 \end{tabular}}
\label{tab: comparison CXR}
\end{table*}
On the other hand, a comparison is conducted with leading existing methods while training on CT images. The results are collected from  \cite{bougourzi2024covid}. Tables \ref{tab:val ct} and \ref{tab:test ct} show these results for the CT dataset for both the validation CT set and the test CT set, respectively. 

\begin{table}
    \caption{Comprehensive Performance Evaluation of the Proposed Method vs. State-of-the-Art Techniques on the Per-COVID-19 validation set \cite{bougourzi2024covid}.}
    \centering    \begin{tabular}{|l|c|c|}\hline
         \textbf{Team}&  \textbf{MAE $\downarrow$}& \textbf{PC $\uparrow$}\\ \hline \hline
\textit{ACVLab}& 4.99&0.9364\\ \hline 
         \textit{EIDOSlab\_Unito}&  4.91& 0.9429\\ \hline 
 \textbf{QCross-Att-PVT}& \textbf{5.42}&\textbf{0.9432}\\ \hline 
 \textit{ausilianapoli94}& 4.95&0.9435\\ \hline 
         \textit{TAC}&   4.48& 0.9460\\ \hline 
         \textit{SenticLab.UAIC}&  4.17& 0.9487\\\hline
 \textit{Taiyuan\_university\_lab713}& 4.50&0.9490\\ \hline
    \end{tabular}
    \label{tab:val ct}
\end{table}
\begin{table}
    \centering
    \caption{Comprehensive Performance Evaluation of the Proposed Method vs. State-of-the-Art Techniques on the Per-COVID-19 test set \cite{bougourzi2024covid}.} \begin{tabular}{|l|c|c|} \hline
         \textbf{Team}&  \textbf{MAE $\downarrow$}& \textbf{PC $\uparrow$}\\ \hline \hline 
 IPLab& 6.53&0.7091\\ \hline  
 \textit{ACVLab}& 4.86&0.7287\\ \hline  
 \textit{SenticLab.UAIC}& 4.61&0.7634\\ \hline 
 \textit{EIDOSlab\_Unito}& 5.02&0.7977\\ \hline 
         \textit{TAC}&   3.64& 0.8022\\ \hline 
 \textbf{QCross-Att-PVT}& \textbf{4.45}&\textbf{0.8094}\\ \hline  
 \textit{Taiyuan\_university\_lab713}& 3.55&0.8547\\ \hline 
    \end{tabular}
    \label{tab:test ct}
\end{table}

\textcolor{black}{The performance of the deep neural networks is assessed by computing the Mean Absolute Error (MAE), the Pearson correlation coefficient (PC), and the standard deviation of absolute error (AE-SD) between the predicted scores and the ground truth values provided by expert radiologists. A perfect MAE is 0, which would indicate completely accurate predictions. The PC measures the strength and direction of the correlation, ranging from -1 to +1, where 0 indicates no correlation, and +1 represents a perfect linear relationship. To provide additional insight into prediction consistency across individual test samples, we include AE-SD, which reflects the spread of error magnitudes for each model’s predictions on the test set.}
\begin{figure}
    \centering
    \includegraphics[width=0.45\textwidth]{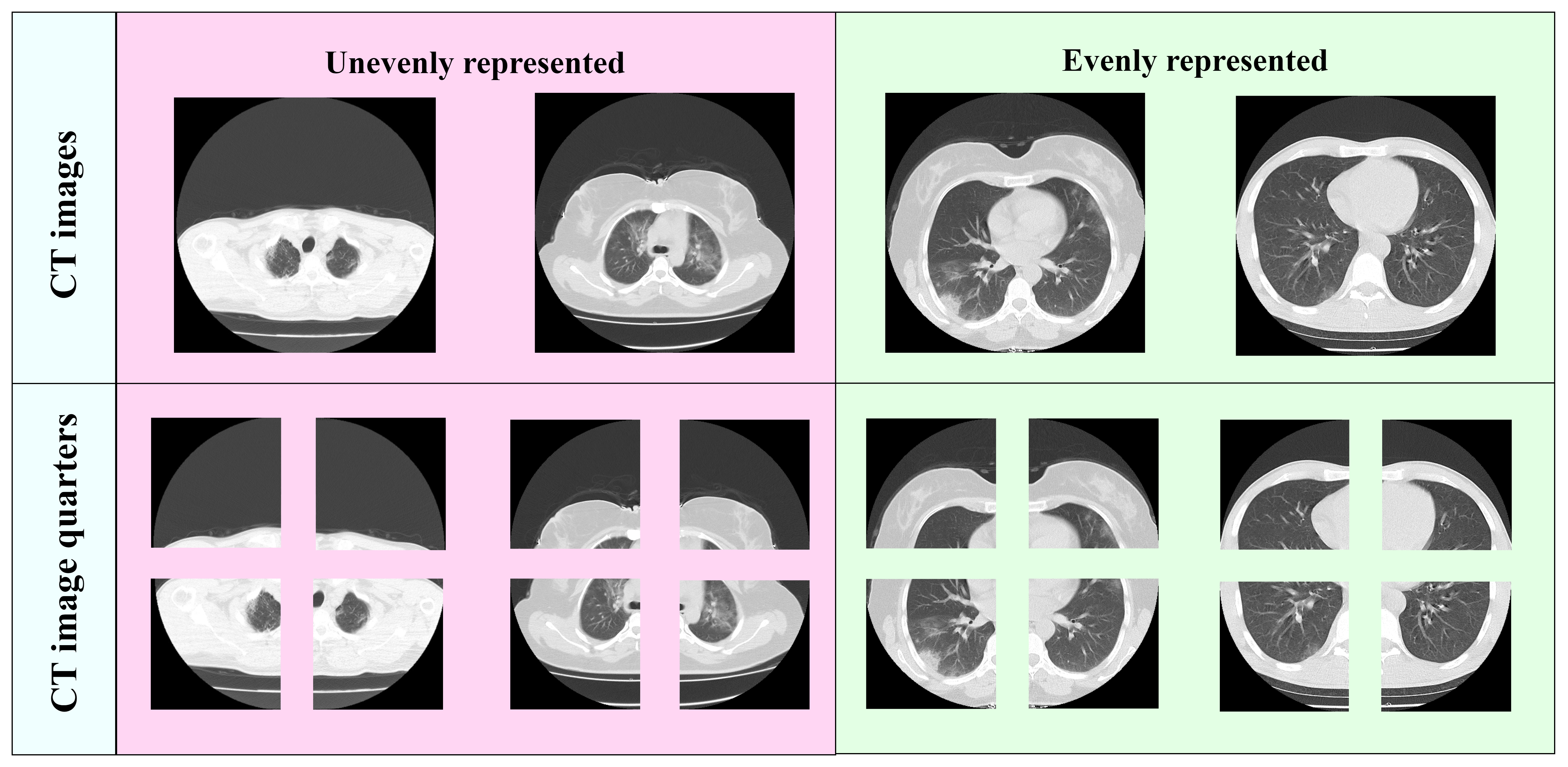}
    \caption{Sample of CT scans of different natures and their corresponding split.}
    \label{fig: CT splits}
\end{figure}

\subsection{Model Interpretability}
\begin{figure*}
    \centering
    \includegraphics[width=0.3\linewidth]{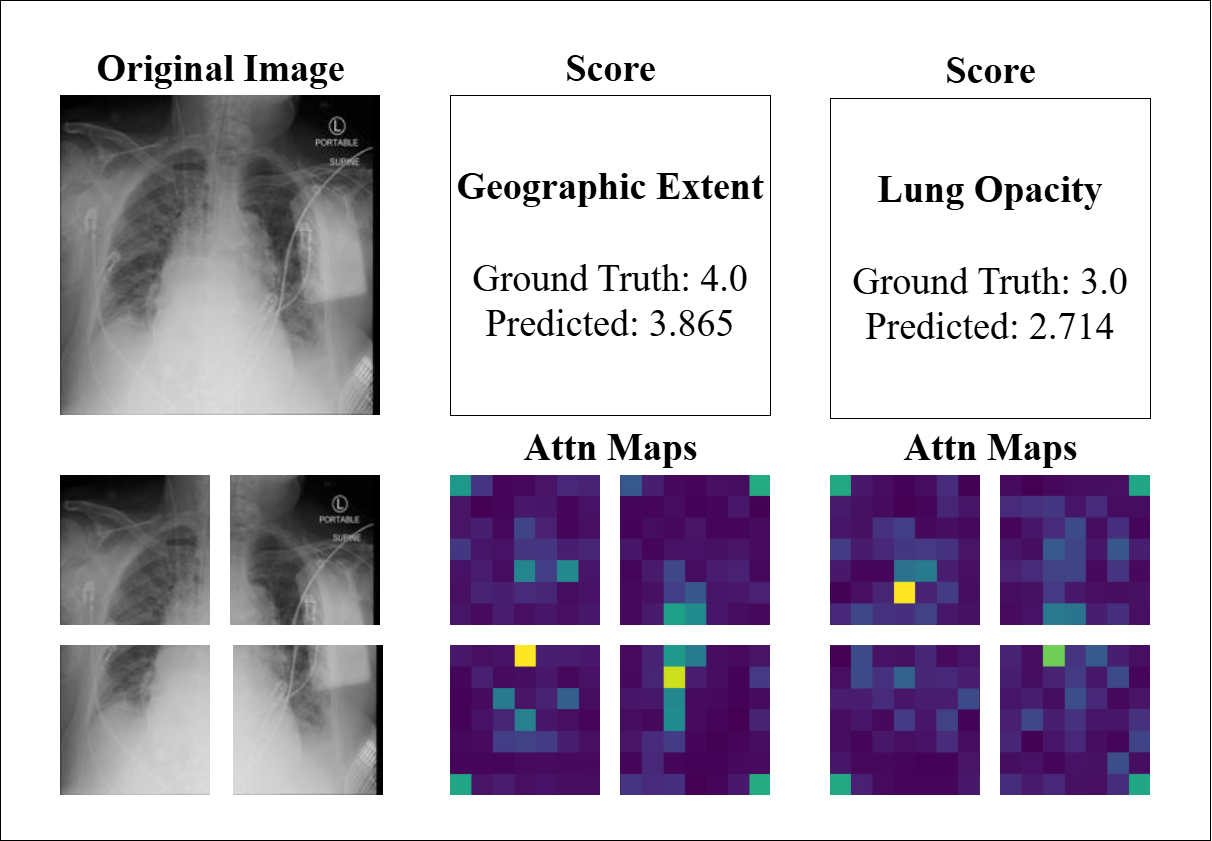}
    \quad
    \includegraphics[width=0.3\linewidth]{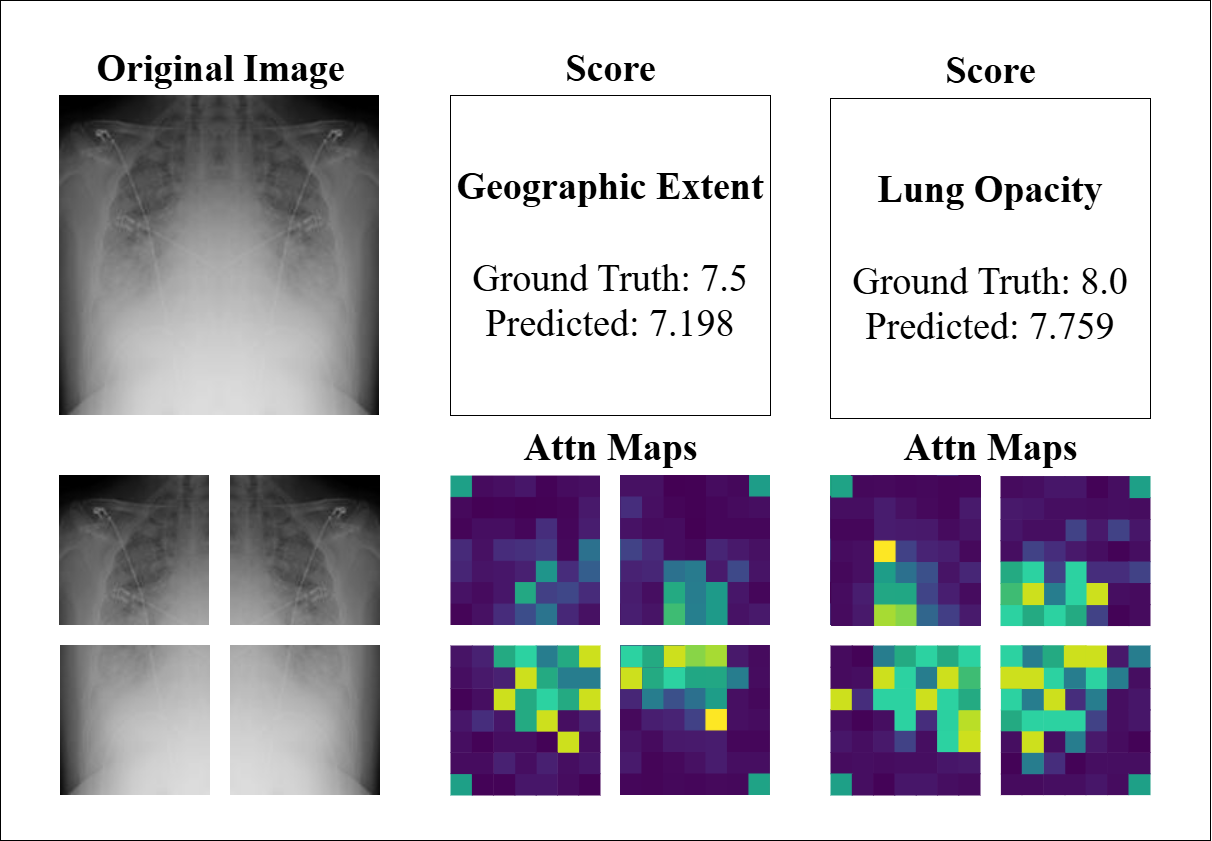}
    \quad
    \includegraphics[width=0.3\linewidth]{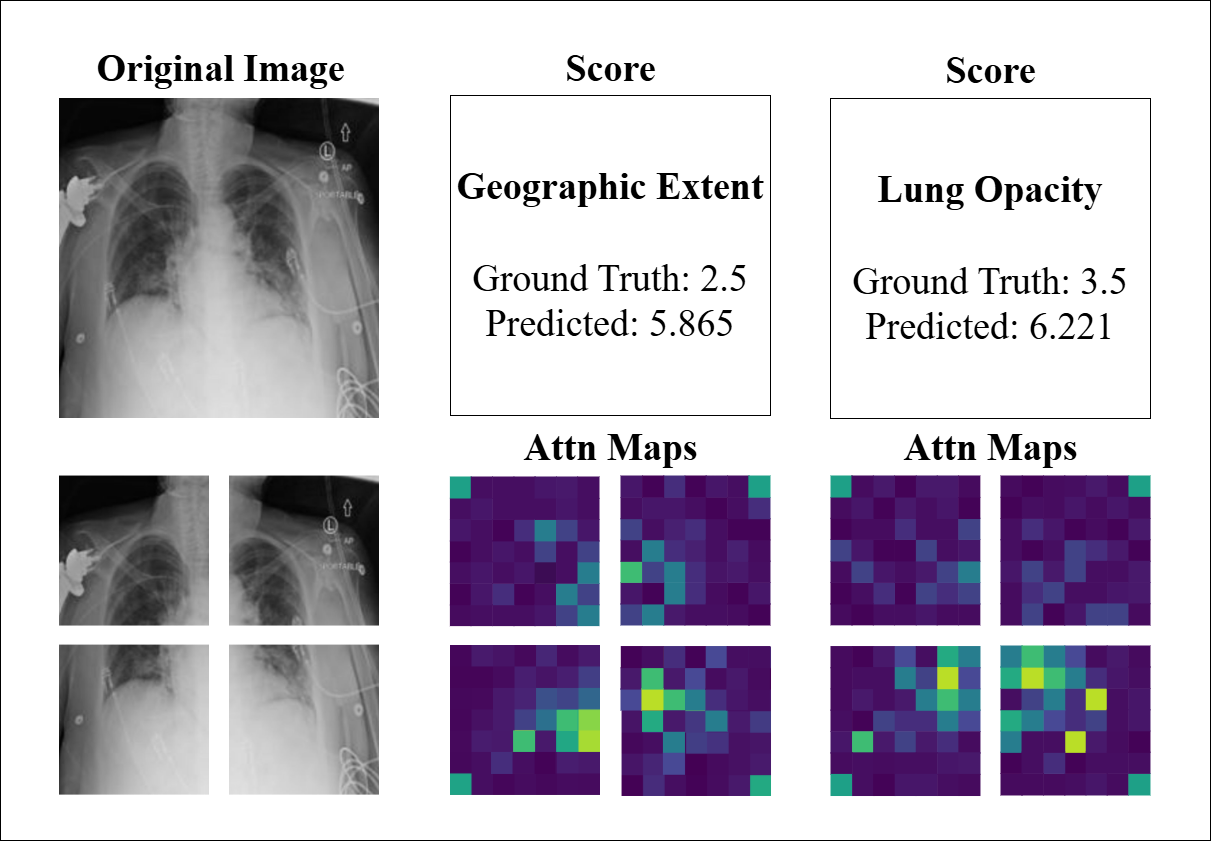}
    \caption{\textcolor{black}{Examples of CXRs with their respective Attention Maps, GT scores and Predictions.}}
    \label{CXR_attn_maps}
\end{figure*}

\textcolor{black}{To further evaluate the interpretability and consistency of our model, we conducted a visual inspection of attention maps and prediction outcomes across representative CXR and CT cases. Figure \ref{CXR_attn_maps} displays three CXR examples alongside their corresponding quadrant-wise attention maps that represent the output of the four PVT encoders, as well as the GT and predicted scores for both GE and LO. The attention maps demonstrate the model’s ability to localize clinically relevant radiographic features, with predictions largely consistent with GT scores (e.g., LO: GT 3.0 vs. predicted 2.714). Notably, the attention weights align with the predicted severity, as the model’s scores reflect the proportion of high-weight regions within the maps. However, the last example (GT 2.5 vs. predicted 5.86) exhibits a significant deviation. While the attention maps highlight diffuse areas of activation, the overestimation may arise from image-quality limitations, such as low contrast or anatomical noise, which could lead the model to misinterpret benign variations as pathological findings. This underscores the influence of technical artifacts on model performance and highlights the need for robust preprocessing to enhance reliability.}

\begin{figure*}
    \centering
     \includegraphics[width=0.2\linewidth]{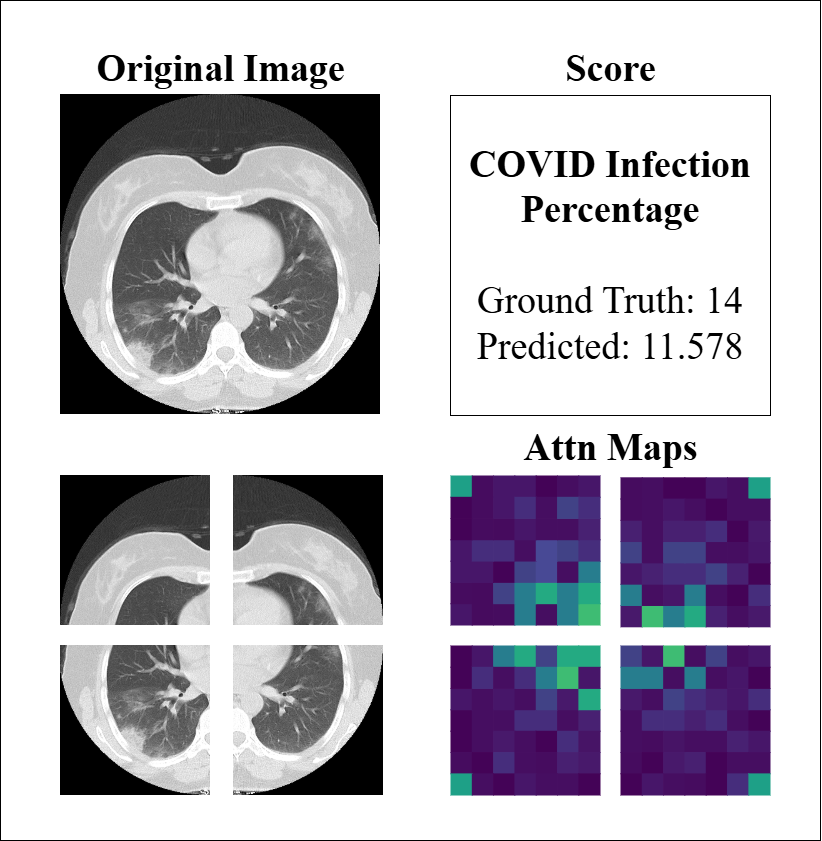}
     \quad 
     \includegraphics[width=0.2\linewidth]{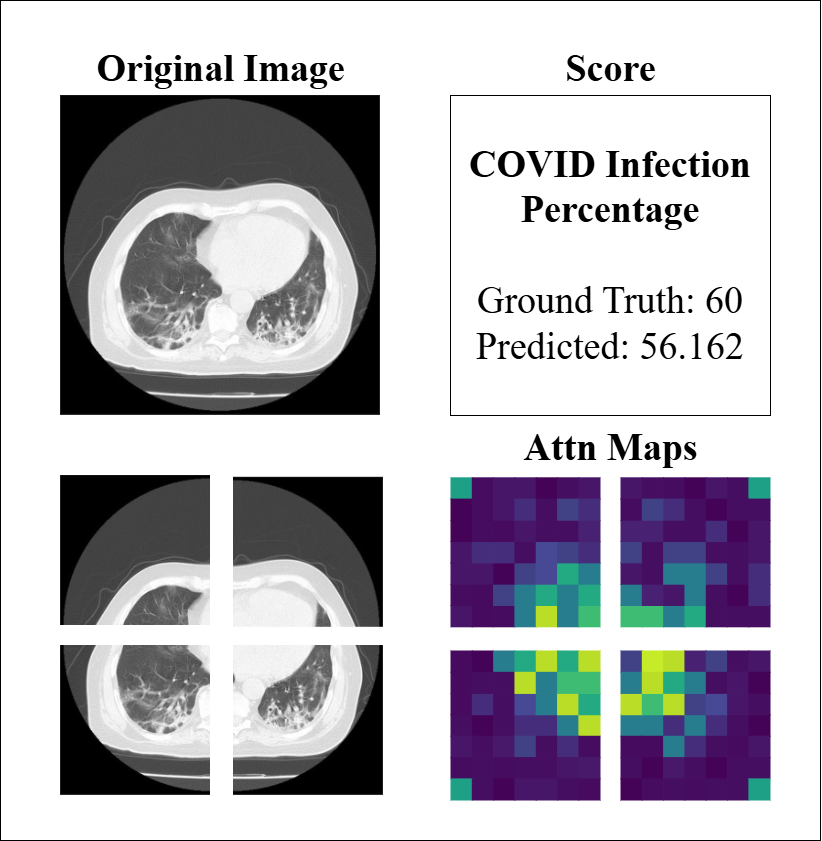}
     \quad
     \includegraphics[width=0.2\linewidth]{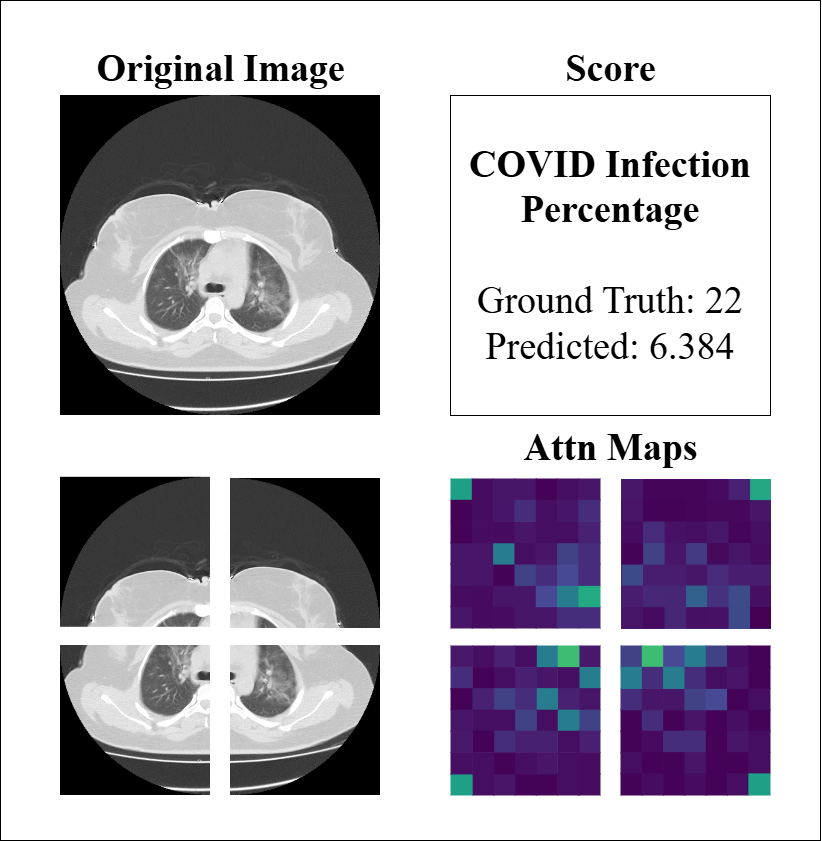}
    \caption{\textcolor{black}{Examples of CT scans with their respective Attention Maps, GT scores and Predictions.}}
    \label{CT_attn_maps}
\end{figure*}

\textcolor{black}{Similarly, to assess the model's performance on CT imaging, we analyzed attention maps and prediction scores across representative cases, as illustrated in Figure \ref{CT_attn_maps}. The examples demonstrate the model's capacity to identify regions of COVID-19 involvement, with attention maps highlighting ground-glass opacities and consolidations that correlate well with infection percentage predictions in most cases (e.g., GT 60 vs. predicted 56.2). The spatial distribution of attention weights aligns with the predicted severity scores, reflecting the model's ability to quantify disease burden through learned feature importance. However, the third example (GT 22 vs. predicted 6.4) reveals a substantial underestimation, despite attention maps showing patchy activations. This discrepancy may be attributed to CT-specific challenges such as partial volume effects, slice thickness variability, or subtle early-stage findings that complicate severity assessment. These observations emphasize that while the model effectively captures pronounced disease patterns, its performance may be limited by technical factors inherent to CT acquisition and early/mild disease manifestations.}


\subsection{Ablation Studies}
This section analyzes the importance of each component of our model through various ablation studies. We first evaluate how the augmentation strategy affects the model's performance. Additionally, we demonstrate the significance of using a custom loss function and its role in enhancing the results. We also test the impact of using a ViT aggregator and examine how the cross-gated attention mechanism affects performance. Additionally, we tested the effect of splitting the image into different numbers of sub-images, thus varying the number of parallel branches. Furthermore, we introduce an ensemble model that further improves results by averaging the predicted scores. These studies are designed to validate the model's robustness across different scenarios and provide deeper insights into its operational dynamics. 

To evaluate the significance of the applied Conditional TransMix as an online augmentation method, we test our proposed QCross-Att-PVT model, both with and without the application of TransMix. The goal of these experiments is to assess how the augmentation strategy affects the model's performance across different tasks and datasets, specifically in predicting GE and LO from CXRs, as well as the CIP scores from CT images. Table \ref{tab: augmentation} presents the results, including the MAE and PC metrics.

\begin{table*}
\centering
\caption{Evaluation of the TransMix Augmentation's Impact on the Proposed Method Using the RALO and Per-COVID-19 Test Sets. }
\begin{tabular}{| l  |l  |l  |l  |l  |l  |l  |l  |l |}\hline
 & \multicolumn{4}{|c|}{\textbf{RALO CXR dataset}}& \multicolumn{4}{|c|}{\textbf{Per-COVID-19 CT dataset}}\\\hline \hline 
 & \multicolumn{2}{|c|}{\textbf{GE}}& \multicolumn{2}{|c|}{\textbf{LO}
}& \multicolumn{2}{|c|}{\textbf{CT val}}& \multicolumn{2}{|c|}{\textbf{CT test}}\\ \hline \hline
 \textbf{TransMix}& \textbf{MAE $\downarrow$}& \textbf{PC $\uparrow$}& \textbf{MAE $\downarrow$}& \textbf{PC $\uparrow$}& \textbf{MAE $\downarrow$}& \textbf{PC $\uparrow$}& \textbf{MAE $\downarrow$}&\textbf{PC $\uparrow$}\\ \hline 
\multicolumn{1}{|c|}{$\times$}& 0.370 & 0.965 & 0.342 & 0.942 & 5.462 & 0.9412 & 4.467 & 0.8081 \\
\hline

\multicolumn{1}{|c|}{$\checkmark$} & \textbf{0.362} & \textbf{0.971} & \textbf{0.337} & \textbf{0.945} & \textbf{5.421} & \textbf{0.9432} & \textbf{4.453} & \textbf{0.8094} \\
\hline

\end{tabular}
    \label{tab: augmentation}
\end{table*}

Similarly, we show the effect of incorporating a weighted loss function on the performance of our proposed QCross-Att-PVT model. We perform experiments with weighted and unweighted loss function. These experiments aim to explore how the use of a weighted loss function impacts the model's performance in predicting GE and LO scores from CXRs, as well as CIP cores from CT images. By applying different weights to the loss associated with various severity scores, we seek to determine whether this strategy improves the model's performance, especially in cases where certain scores are less represented in the training dataset. The outcomes of these experiments, measured by MAE and PC, are detailed in Table \ref{tab: loss}. 

\begin{table*}
\centering
\caption{Evaluation of the Loss Function's Effectiveness on the Proposed Method Using the RALO and Per-COVID-19 Test Sets. }
\begin{tabular}{| l  |l  |l  |l  |l  |l  |l  |l  |l |} \hline 
 & \multicolumn{4}{|c|}{\textbf{RALO CXR dataset}}& \multicolumn{4}{|c|}{\textbf{Per-COVID-19 CT dataset}}\\\hline \hline 
 & \multicolumn{2}{|c|}{\textbf{GE}}& \multicolumn{2}{|c|}{\textbf{LO}
}& \multicolumn{2}{|c|}{\textbf{CT val}}& \multicolumn{2}{|c|}{\textbf{CT test}}\\ \hline \hline
 \textbf{Loss Function}& \textbf{MAE $\downarrow$}& \textbf{PC $\uparrow$}& \textbf{MAE $\downarrow$}& \textbf{PC $\uparrow$}& \textbf{MAE $\downarrow$}& \textbf{PC $\uparrow$}& \textbf{MAE $\downarrow$}&\textbf{PC $\uparrow$}\\ \hline 
$L_1$ & 0.365& 0.969& 0.341& 0.944& 5.501& 0.9392& 4.499& 0.8011\\
\hline
Weighted & \textbf{0.362} & \textbf{0.971} & \textbf{0.337} & \textbf{0.945} & \textbf{5.421} & \textbf{0.9432} & \textbf{4.453} & \textbf{0.8094} \\
\hline

\end{tabular}
    \label{tab: loss}
\end{table*}
To evaluate the impact of using a ViT as a feature aggregator in our model, we conduct experiments that compare the performance of the model with and without this component. Specifically, we compare the proposed QCross-Att PVT model, which includes a ViT aggregator, with a modified version of the same model where the ViT has been removed and replaced with a global average pooling (GAP) block. This approach reduces the dimensions by averaging the values across each feature map, resulting in a single feature vector. This vector is then fed directly into the regression head, bypassing the need for the ViT to further process the aggregated features. The results of this ablation study is summarized in Table \ref{tab: vit test}, which reports the performance metrics, including MAE and PC. By comparing these metrics, we can determine the effectiveness of the ViT as a feature aggregator in enhancing the model's ability to predict severity scores accurately.

 \begin{table*}
\centering
\caption{Evaluation of the Impact of  ViT Aggregator on the Proposed Method Using the RALO and Per-COVID-19 Test Sets.}
\begin{tabular}{| l  |l  |l  |l  |l  |l  |l  |l  |l |} \hline 
 & \multicolumn{4}{|c|}{\textbf{RALO CXR dataset}}& \multicolumn{4}{|c|}{\textbf{Per-COVID-19 CT dataset}}\\\hline \hline 
 & \multicolumn{2}{|c|}{\textbf{GE}}& \multicolumn{2}{|c|}{\textbf{LO}
}& \multicolumn{2}{|c|}{\textbf{CT val}}& \multicolumn{2}{|c|}{\textbf{CT test}}\\ \hline \hline
 \textbf{Aggregator}& \textbf{MAE $\downarrow$}& \textbf{PC $\uparrow$}& \textbf{MAE $\downarrow$}& \textbf{PC $\uparrow$}& \textbf{MAE $\downarrow$}& \textbf{PC $\uparrow$}& \textbf{MAE $\downarrow$}&\textbf{PC $\uparrow$}\\ \hline 
GAP& 0.367& 0.965& 0.381& 0.942& 5.498& 0.9295& 4.501& 0.7897\\
\hline

ViT& \textbf{0.362} & \textbf{0.971} & \textbf{0.337} & \textbf{0.945} & \textbf{5.421} & \textbf{0.9432} & \textbf{4.453} & \textbf{0.8094} \\
\hline

\end{tabular}
    \label{tab: vit test}
\end{table*}

By eliminating the cross-attention mechanism, the model’s ability to capture relationships and dependencies between different regions of the lung images may be diminished, potentially affecting the accuracy of the severity predictions. To evaluate the impact of incorporating gated cross-attention within the encoder, we compared the model's performance with and without this mechanism. Specifically, we tested the proposed QCross-Att-PVT model, which integrates cross-attention modules in its encoder, against a modified version where these components were excluded. In the modified model, the encoder processes each of the four image regions independently, and the resulting tensors are directly concatenated. The outcomes of these experiments, including MAE and PC, are shown in Table \ref{tab: cross attn}.
 \begin{table*}
\centering
\caption{Evaluation of the Impact of Gated Cross-Attention on the Proposed Method Using the RALO and Per-COVID-19 Test Sets.}
\begin{tabular}{| l  |l  |l  |l  |l  |l  |l  |l  |l |} \hline 
 & \multicolumn{4}{|c|}{\textbf{RALO CXR dataset}}& \multicolumn{4}{|c|}{\textbf{Per-COVID-19 CT dataset}}\\\hline \hline 
 & \multicolumn{2}{|c|}{\textbf{GE}}& \multicolumn{2}{|c|}{\textbf{LO}
}& \multicolumn{2}{|c|}{\textbf{CT val}}& \multicolumn{2}{|c|}{\textbf{CT test}}\\ \hline \hline
 \textbf{Gated cross-attention}& \textbf{MAE $\downarrow$}& \textbf{PC $\uparrow$}& \textbf{MAE $\downarrow$}& \textbf{PC $\uparrow$}& \textbf{MAE $\downarrow$}& \textbf{PC $\uparrow$}& \textbf{MAE $\downarrow$}&\textbf{PC $\uparrow$}\\ \hline 
\multicolumn{1}{|c|}{$\times$}& 0.371& 0.961& 0.342& 0.938& 5.512& 0.9325& 4.511& 0.7684\\
\hline

\multicolumn{1}{|c|}{$\checkmark$} & \textbf{0.362} & \textbf{0.971} & \textbf{0.337} & \textbf{0.945} & \textbf{5.421} & \textbf{0.9432} & \textbf{4.453} & \textbf{0.8094} \\
\hline

\end{tabular}
    \label{tab: cross attn}
\end{table*}
\textcolor{black}{To assess the effect of different image partitioning strategies, we conducted a controlled ablation study in which the input images were divided into varying numbers of spatial sub-regions. Specifically, we experimented with configurations where the images were split into 2, 4, and 6 equally sized regions. Each configuration was integrated into our model while keeping all other components and training parameters constant. The experiments were conducted across both imaging modalities, CXR and CT, to ensure the observed trends were not modality-specific. Performance metrics, including MAE and PC, were collected and summarized in Table \ref{tab: split_comparison}.}

\begin{table*}
    \centering
      \caption{\textcolor{black}{Ablation study showing the effect of varying the number of split regions on the performance.}}
\begin{tabular}{|c|c|c|c|c|c|c|c|c|}
        \hline
        & \multicolumn{4}{|c|}{\textbf{CXR}} & \multicolumn{4}{|c|}{\textbf{CT}}\\ \hline \hline
 & \multicolumn{2}{|c|}{\textbf{GE}}& \multicolumn{2}{|c|}{\textbf{LO}}& \multicolumn{2}{|c|}{\textbf{CT val}} & \multicolumn{2}{|c|}{\textbf{CT test}}\\
        \hline \hline
        \textbf{Regions}& \textbf{MAE $\downarrow$} & \textbf{PC} & \textbf{MAE $\downarrow$} & \textbf{PC} & \textbf{MAE $\downarrow$} & \textbf{PC}  & \textbf{MAE $\downarrow$} &\textbf{PC} 
\\
        \hline
        2& 0.459& 0.916& 0.430& 0.886& 5.984&  0.9321& 5.008&
0.7879\\
        \hline
        \textbf{4}& \textbf{0.362}&\textbf{0.971}&\textbf{ 0.337}& \textbf{0.945}& \textbf{5.421}& \textbf{0.9432}& \textbf{4.453}&
\textbf{0.8094}\\
        \hline
        6& 0.437& 0.922& 0.416& 0.901& 5.784&  0.9358& 4.893&0.7795\\ \hline
\end{tabular}
 
    \label{tab: split_comparison}
\end{table*}

We propose an ensemble model that combines the strengths of the best-performing architectures to enhance prediction accuracy. The core idea is to leverage multiple models by averaging their outputs to produce a final prediction. Each of the selected models, which demonstrated superior performance in individual tasks, generates its predicted scores. These individual predictions are then averaged to form a consolidated score, which is subsequently compared against the ground truth scores. By aggregating the outputs from multiple architectures, the ensemble model aims to reduce the impact of errors from any single model, leading to more robust and accurate predictions across different tasks and datasets.
The individual models used in the ensemble are variations of the proposed QCross-Att-PVT model, where either the encoders or the aggregator transformer are modified. We present three specific models: the first, which is presented in this paper, utilizes a PVT encoder with a ViT aggregator, the second employs a ViT encoder with a PVT aggregator, and the third combines both PVT encoders and a PVT aggregator. By averaging the outputs of these three models, the ensemble aims to enhance the overall prediction accuracy by leveraging the strengths of each configuration while mitigating their individual weaknesses. Table \ref{tab: ensemble} shows the results of each of the three models and the ensemble method in terms of MAE and PC.
 
\begin{table*}
\centering
\caption{Performance evaluation of modified versions of the proposed model, in addition to the ensemble model.}

\begin{tabular}{|l |l  |l  |l  |l  |l  |l  |l  |l  |l |} \hline 
 \multicolumn{2}{|c|}{}& \multicolumn{4}{|c|}{\textbf{RALO CXR dataset}}& \multicolumn{4}{|c|}{\textbf{Per-COVID-19 CT dataset}}\\ \hline \hline
 \multicolumn{2}{|c|}{\textbf{Model}}& \multicolumn{2}{|c|}{\textbf{GE}}& \multicolumn{2}{|c|}{\textbf{LO}
}& \multicolumn{2}{|c|}{\textbf{CT val}}& \multicolumn{2}{|c|}{\textbf{CT test}}\\ \hline \hline
 \textbf{Encoder}& \textbf{Aggregator}& \textbf{MAE $\downarrow$}& \textbf{PC $\uparrow$}& \textbf{MAE $\downarrow$}& \textbf{PC $\uparrow$}& \textbf{MAE $\downarrow$}& \textbf{PC $\uparrow$}& \textbf{MAE $\downarrow$}&\textbf{PC $\uparrow$}\\ \hline 
 ViT&PVT& 0.399& 0.955& 0.328& 0.930& 5.623& 0.9291& 4.521& 0.7772\\
\hline
 PVT&PVT& 0.385& 0.959& 0.342& 0.934& 5.642& 0.9328& 4.501& 0.7799\\
\hline
 PVT&VIT& 0.362 & 0.971 & 0.337 & 0.945 & 5.421 & 0.9432 & 4.453 & 0.8094 \\
\hline\hline
 \multicolumn{2}{|c|}{\textbf{Ensemble Model}}& 0.358& 0.973& 0.336& 0.948& 5.409& 0.9444& 4.439&0.8107\\\hline

\end{tabular}
    \label{tab: ensemble}
\end{table*}

\section{Discussion}
\label{Discussion}
Table \ref{tab: comparison CXR} shows that our proposed QCross-Att-PVT model achieves the lowest MAE of \textbf{0.362} for GE and \textbf{0.337} for LO, indicating that our model provides the most accurate predictions. Moreover, it also achieves the highest PC values of \textbf{0.971} for GE and \textbf{0.945} for LO, showing a very strong correlation between the predicted scores and the actual ground truth values. \textcolor{black}{Furthermore, the remarkably low AE-SD values (\textbf{0.0453} for GE and \textbf{0.0478} for LO) demonstrate the model’s ability to produce stable and reliable predictions across different patient cases.} These results highlight the superior predictive capability and reliability of our proposed approach. In addition to the comparison with state-of-the-art approaches on the RALO CXR dataset, Table \ref{tab: comparison CXR} also compares the number of parameters of each model, which provides valuable insight into the complexity and computational requirements of the models. Our model has a comparatively higher number of parameters, offering higher performance at the cost of higher processing time and memory requirements. This demonstrates that a higher number of parameters can be crucial to achieving the best results in complex medical imaging tasks.

\textcolor{black}{To validate the robustness of our proposed framework, we performed five independent training runs during development, using different random seeds and identical settings. Across these runs, we consistently observed very similar performance trends, with low standard deviations for both MAE and PC. Specifically, the  MAE standard deviation was \textbf{0.0032} for GE and \textbf{0.0051} for LO, while the standard deviation of the Pearson correlation was \textbf{0.0009} for GE and \textbf{0.0018} for LO. These small values indicate that the model’s performance is stable and reliable across different training instances, demonstrating low sensitivity to initialization and stochastic optimization effects. While we report these statistics to illustrate the consistency of our approach, we note that providing comparable multi-run standard deviations for all baseline methods was not feasible, as many existing SOTA models were reproduced only once using publicly available configurations. For this reason, our main tables focus on mean performance metrics while the additional repeated-run analysis supports the robustness of our model’s training dynamics.}

On the other hand, Tables \ref{tab:val ct} and \ref{tab:test ct} reveal that our proposed model has a comparative performance against several state-of-the-art models on the Per-COVID-19 dataset, both for the validation and test sets. As shown in Table \ref{tab:val ct}, our model achieves an MAE of \textbf{5.421} (the scale range is 100) and a PC of \textbf{0.9432} for the validation set. Although this MAE is slightly higher than that of the top-performing models such as ``Taiyuan\_university\_lab713" (MAE of \textbf{4.50}), our model's performance in terms of PC metric is still competitive, reflecting a high degree of correlation between the predicted and ground truth values. Notably, our PC value of \textbf{0.9432} is comparable to that of the leading teams, such as ``ausilianapoli94" (PC of \textbf{0.9435}). This indicates that our model is accurate and its prediction consistency is strong. Similarly, this is shown in Table \ref{tab:test ct} where the results over the test set are presented. Here, our model exhibits a more competitive performance with an MAE of \textbf{4.45}, and a PC of \textbf{0.8094}. This MAE value is lower than those achieved by several other models and our PC value of \textbf{0.8094} is higher than those of many competing models, like ``SenticLab.UAIC" (PC of \textbf{0.7634}), reflecting a better correlation between the predicted scores and the actual values. The differences in MAE and PC values between our model and the top performers are marginal, especially considering the scores range (0 to 100). 

To compare performance more effectively in terms of generalization ability, it's important to highlight that the validation labels are naturally noisy. This is because they are based on radiologists' direct estimation of the infection percentage by visually comparing the infected areas to the total lung volume. However, despite the difficulty of learning from such noisy data, our approach shows a strong capacity to generalize in domain adaptation scenarios. 
Moreover, concerning our method for dividing CT images into regions of interest, we recognize that in many CT slices, the division into four lung quarters is not always uniform and may not consistently represent the complete anatomical structure as shown in Figure \ref{fig: CT splits}. 

Despite these potential limitations in precisely dividing the lungs into uniform quarters, our model still performs competitively compared to the leading approaches. For future work, we aim to enhance this splitting process by employing more advanced techniques, such as dedicated lung detection or segmentation methods, to achieve greater consistency and accuracy in dividing the lungs into quarters. By incorporating these refined approaches, we expect to improve the anatomical relevance of the segmentation, leading to even better model performance in capturing the complex patterns present in the lung images. 

Overall, these tables illustrate that our model demonstrates robust performance, achieving relatively low MAE and high PC values compared to the state-of-the-art models. Its ability to maintain strong correlation scores indicates its reliability and effectiveness in severity prediction tasks, for both CXR and CT image analysis.

The ablation studies presented in Table \ref{tab: augmentation} demonstrate the clear benefits of applying Conditional TransMix as an online augmentation method to our QCross-Att-PVT model. The results consistently show that the model performs better with TransMix across various datasets and prediction tasks. This improvement is reflected in the model's ability to generate predictions that are more closely aligned with the ground truth, highlighting its enhanced generalization capabilities. These findings demonstrated by a lower MAE and a higher PC confirm that the augmentation method significantly boosts the model's learning efficiency and overall performance.

Similarly, our experiments reveal the positive effects of using a weighted loss function, as shown in Table \ref{tab: loss}. The results suggest that incorporating a weighted loss function enhances the model's performance by allowing it to assign different importance to underrepresented scores in the training data. This strategy enables the model to capture less frequent but clinically significant cases more accurately, thereby reducing prediction errors and improving the correlation between predicted scores and actual clinical outcomes.

Further, the results in Table \ref{tab: vit test} illustrate the substantial impact of using a ViT as a feature aggregator within the QCross-Att-PVT model by resulting in a lower MAE and higher PC. The findings indicate that the ViT is more effective in capturing complex feature representations, which enhances the model's predictive capabilities. In contrast, the global average pooling (GAP) method, which simplifies feature maps, may lose important spatial and contextual information needed for precise predictions.

The results in Table\ref{tab: cross attn} show that the inclusion of gated cross-attention significantly improves the performance of the model for both the RALO CXR and the Per-COVID-19 CT datasets. For the RALO dataset, the inclusion of gated cross-attention leads to a lower MAE and a higher PC, indicating improved prediction accuracy and a stronger correlation with ground truth. For the per-COVID-19 CT dataset, the model also achieves a lower MAE and a higher PC for both the validation and test set when this mechanism is applied. These improvements highlight the effectiveness of gated cross-attention in capturing dependencies between different image regions, leading to more accurate and reliable predictions. By facilitating the exchange of information between different regions of the input image, gated cross-attention enables the model to better understand complex patterns and relationships.

\textcolor{black}{The results of the ablation study in Table \ref{tab: split_comparison}, reveal that splitting the input image into four quadrants consistently yielded the best performance across all evaluation metrics and imaging modalities. When using only two regions, the model struggled to capture sufficient localized detail, resulting in a reduced ability to model spatial variation in lung abnormalities. On the other hand, increasing the number of regions to six introduced smaller, overly fragmented sub-images, which likely impaired semantic coherence and led to diminishing returns in performance despite increased complexity. The four-region configuration offers a favorable balance between anatomical relevance, representational capacity, and computational efficiency. Notably, it aligns well with the structure of PVT backbones, which are optimized for processing square-shaped patches, thus reinforcing the design choice in our architecture.}

For the ensemble model, Table \ref{tab: ensemble} highlights that while each individual model configuration has unique strengths, the third model (PVT encoder and ViT aggregator) generally outperforms others in terms of MAE and PC across most datasets. However, the ensemble model, which combines the outputs of all three variations, achieves the best overall performance. It consistently delivers the lowest MAE and highest PC values for both GE and LO predictions on the RALO CXR dataset, as well as for CIP scores on the Per-COVID-19 CT validation and test sets. These findings emphasize the effectiveness of an ensemble approach, as it leverages the strengths of multiple models, mitigates individual model weaknesses, and reduces the impact of errors from any single model. The ensemble model's slight improvements in MAE and PC values demonstrate its superior generalization ability and resilience to data variability, making it a powerful strategy for enhancing performance in complex medical imaging tasks.

The ablation studies confirm that the selected parameters consistently lead to improved performance in both modalities. This validates our choice of parameters,  that our model with its specific configuration achieves high performance in predicting severity scores for both CXRs and CT scans. These results emphasize the effectiveness of the detailed structure of our model in achieving excellent results across different medical imaging tasks.

\textcolor{black}{While our proposed model demonstrates strong performance across two medical imaging modalities, several limitations should be noted in the context of clinical deployment. First, the relatively high number of parameters (28M) may pose challenges for real-time inference on limited hardware, such as edge devices in resource-constrained settings. Second, our method has been validated on two COVID-19-specific datasets, despite the common characteristics between COVID-19 and other pneumonia infections \cite{sun2020systematic, bougourzi2023pdatt}. The generalizability to other types of pulmonary infections or non-COVID conditions warrants further exploration. Finally, integration into clinical workflows would benefit from interactive interfaces that visualize model outputs alongside radiological findings, enabling clinicians to validate or contest predictions. Future work will address these limitations through model compression, cross-institutional validation, and clinician-in-the-loop frameworks.}
 
\textcolor{black}{Although the datasets used in this study are specific to COVID-19-related infections, they remain among the few publicly available resources offering continuous, radiologist-assessed severity scores. This makes them particularly valuable for evaluating regression-based severity models. In contrast, many public datasets are limited to classification tasks and do not support direct severity estimation. Our use of RALO and Per-COVID-19 thus allows for fair benchmarking with prior studies. Moving forward, we aim to extend our approach to broader pulmonary diseases by incorporating more diverse datasets and applying domain adaptation to enhance generalizability across different clinical scenarios.}
\section{Conclusion}
\label{conclusion}
Accurate identification of pneumonia is critical to optimize patient care, prevent disease transmission, and initiate public health interventions to reduce the impact of this widespread and potentially serious respiratory disease. Our study presents a novel approach that combines a parallel design of PVTs with cross-gated attention as an encoder and feature processing with a ViT to improve the computational assessment of lung disease severity. The results show that our model performs superiorly when a weighted loss function is applied. The results across multiple modalities outperform various deep learning models and several state-of-the-art techniques. In particular, our method accurately quantifies the severity of lung disease when applied to CXRs and CT scans, providing reliable predictions that allow physicians to make more direct and objective assessments. In addition, the application of a conditional online augmentation technique helps to balance
 the training data and further improve the prediction accuracy. Overall, our proposed approach not only improves the accuracy of automatic lung severity assessment but also provides physicians with a versatile tool for diagnosis and treatment planning.


\end{document}